\documentclass[cameraready]{Interspeech}
\title{Towards Digital Preservation of Efik: TTS for a Low-Resource African Language}

\author[affiliation={1,3}, correspondingauthor]{Offiong Bassey}{Edet}
\author[affiliation={1}]{Emmanuel}{Oyo-Ita}
\author[affiliation={2}]{Archibong Okon}{Archibong}
\author[affiliation={2}]{David Effanga}{Bassey}
\author[affiliation={3}]{Mbuotidem Sunday}{Awak}
\address{
    $^1$ University of Cross River State, Nigeria \\
    $^2$ University of Calabar, Nigeria \\
    $^3$ ML Collective
}

\email{offiongbassey99@gmail.com, emmanueloyoita@unicross.edu.ng, archibongarchibong54@gmail.com, daviba231@gmail.com, mbuotidemawak@gmail.com}

\keywords{text-to-speech, low-resource languages, Efik language, tonal languages, speech synthesis, African languages}

\usepackage{comment}

\begin{document}

\maketitle

\begin{abstract}
   Efik, a tonal language spoken by about 3 million second language speakers and 1.5 million native speakers in Southeastern Nigeria, remains underrepresented in speech synthesis research. We present the first documented end-to-end text-to-speech study for Efik, introducing a curated single speaker corpus of 2,632 utterances totaling three hours and a comparative evaluation of four neural models (VITS, MMS-TTS, SpeechT5, and Orpheus-TTS) under low resource conditions. Native speakers evaluated the systems using MOS, Nat-MOS, and A-MOS. MMS-TTS achieved the highest MOS of 3.80 ± 0.63 and produced more stable long form speech, though tonal errors persisted. Other models showed greater tonal and prosodic inconsistencies. These results provide a reproducible baseline and highlight the need for larger corpora and tone aware modeling for tonal African languages.
\end{abstract}

\section{Introduction}

Text-to-Speech (TTS) technology has made remarkable progress in producing speech that approaches human naturalness in high-resource languages \cite{ogayo2022africanvoices, pine2025indigenous}. Modern end-to-end architectures, often combining attention mechanisms with neural vocoders, achieve high-quality synthesis when trained on large-scale paired text–audio corpora. However, these successes rely on hundreds of hours of curated speech data, a requirement that remains a major obstacle for low-resource languages, where such datasets are scarce or entirely unavailable \cite{magueresse2020lowresource, diack2026waxal, kharitonov2023speak}.

Efik, a Lower Cross language spoken in Southeastern Nigeria, exemplifies this challenge. Despite an estimated 3 million second speakers and 1.5 million native speakers \cite{mensah2014efik}, Efik lacks publicly available speech datasets suitable for supervised TTS training and remains largely absent from contemporary speech technology pipelines, reflecting broader patterns of digital language inequality \cite{kornai2013digital}. While text-based NLP tasks such as machine translation for Efik language have been receiving attention, speech systems remain critically underdeveloped \cite{kalejaiye2025ibomnpl, edet2026efikmt}.

Developing TTS for Efik presents several technical challenges. First, severe data scarcity constrains the training of data-intensive neural architectures \cite{katumba2025buildingtts}. Second, Efik is a tonal language in which lexical and grammatical contrasts are encoded through pitch variation \cite{yip2002tone}. Accurate synthesis therefore requires careful modeling of tonal contours in addition to segmental phonetic structure. Inadequate tone realization can alter lexical meaning, reducing intelligibility and perceived naturalness.

To investigate the feasibility of neural speech synthesis under such constraints, we curated a high-quality Efik speech corpus consisting of approximately three hours of single-speaker recordings captured with a wireless microphone in controlled conditions. The dataset comprises 2,632 utterances designed to provide phonetic and tonal coverage for supervised training. Operating within this limited-data regime, we finetune and evaluate four neural TTS models to assess naturalness and intelligibility in synthesized Efik speech.

This research addresses these critical gaps by developing TTS systems specifically tailored for Efik, contributing to the broader effort of digital language preservation and ensuring that speakers of low-resource languages can participate fully in the digital ecosystem.

\section{Efik Language Documentation and Computational Efforts}
Historically, Efik has enjoyed a robust literary status due to early missionary documentation, such as the dictionaries of Hugh Goldie and R.G.F. Adams, and the extensive cultural records of indigenous historians like E. U. Aye. However, these works remained largely “analogue,” restricted to hard copies and physical libraries. While later promoters like Ita \cite{ita2016efikproverbs} expanded this corpus with multi-volume thematic studies on Efik proverbs.  Mensah \cite{mensah2014efik}  notes a simultaneous decline in vitality among younger urban speakers who prioritize English and Nigerian Pidgin.

The 21st century marks a critical transition toward digital preservation to bridge this “digital language divide.” Akoda \cite{akoda2022analogue} establishes the foundational need to migrate deteriorating analogue manuscripts into digital repositories, highlighting early mobile successes like the Learn Efik app and the Tete Efik Dictionary. While these tools improved accessibility, challenges remain regarding tonal accuracy for non-native speakers.

Recent research underscores a critical shift toward digital preservation for Nigeria’s minority languages. While, Akoda \cite{akoda2022analogue} establishes the foundational need to transition Efik cultural and historical records from deteriorating analogue manuscripts into digital repositories, Kalejaiye \cite{kalejaiye2025ibomnpl} demonstrate the technical application of such digitized data through the Ibom NLP project, creating the essential text corpora required for Anang, Efik, Ibibio, and Oro to participate in global AI ecosystems. Complementing these efforts, Ekpenyong \cite{ekpenyong2025multimodal} introduce a multimodal framework using digitized speech and geospatial data to automate the real-time assessment of language vitality. 

Together, these advancements represent a move from passive archiving to active computational linguistics, ensuring Efik remains a functional, “living” language in the age of Artificial Intelligence and also for the next generation of digital speakers.

\section{Related Work}
Recent advances in neural text-to-speech (TTS) have stimulated growing interest in speech technologies for African languages. Several studies have explored TTS for relatively higher-resource African languages, including Yoruba, Igbo, Hausa, Swahili, Akan, Lagunda, Lingala, and Ewe \cite{kagumire2024luganda, ogayo2022africanvoices, meyer2022bibletts, ogunremi2024iroyinspeech, aliero2020crosslingual, mbonimpa2025edge}. Large-scale multilingual efforts such as WAXAL further expanded coverage by providing speech resources for 24 African languages \cite{diack2026waxal}.

Beyond African-language coverage, prior work has examined the broader challenge of low-resource speech synthesis, including data efficiency, minimal-supervision TTS, and speech generation for language revitalization and education \cite{pine2022requirements, pine2025indigenous, kharitonov2023speak}. These studies highlight persistent constraints such as limited recordings, speaker scarcity, orthographic variation, and evaluation challenges in underrepresented languages.

Despite this progress, the Efik language remains severely underrepresented in speech technology research, with no prior systematic benchmark for TTS to the best of our knowledge. Our work addresses this gap by introducing a curated Efik speech corpus and evaluating multiple modern TTS SOTA models.

\section{Dataset Creation}
We curated approximately three hours of high-quality speech data from a single native Efik speaker. Recordings were captured using a wireless microphone in a quiet indoor environment to minimize background noise and ensure acoustic consistency. The final corpus consists of 2,632 utterances segmented at the sentence level.

The recording materials were drawn from Efik novels, folktales, and educational texts in order to provide lexical diversity and broad domain coverage. Care was taken to include varied sentence structures and tonal patterns to support supervised speech synthesis under low-resource conditions.

Although Efik is spoken by 1.5 million native speakers \cite{mensah2014efik}, obtaining a high-quality speech dataset suitable for neural Text-to-Speech training presents significant challenges. Publicly available audio resources often contain background music, environmental noise, or overlapping speech, rendering them unsuitable for controlled acoustic modeling. Furthermore, many existing recordings lack explicit usage consent for computational research and TTS development, limiting their applicability.

These constraints necessitated the collection of a dedicated, consented, and acoustically controlled dataset specifically designed for supervised Efik TTS development.

\subsection{Data Labeling}
We initially explored automatic forced alignment to generate time-aligned transcriptions for the recorded Efik speech data. However, this approach proved unreliable due to the absence of robust Automatic Speech Recognition (ASR) models for Efik. Pretrained multilingual models, including Whisper \cite{radford2023whisper} and XLS-R \cite{babu2021xlsr}, produced highly inaccurate transcriptions when applied to the segmented Efik utterances, likely due to limited exposure to Lower Cross languages during pretraining.

Given the low transcription accuracy and the tonal sensitivity of the language, we opted for manual annotation of the entire corpus. Each of the 2,632 utterances was carefully transcribed and verified to ensure orthographic consistency and alignment with the recorded speech. This manual labeling process ensured high-quality paired text–audio data suitable for supervised TTS training.

\subsection{Dataset Validation}
The curated TTS corpus underwent a manual validation process to ensure transcription accuracy and phonetic fidelity. A native Efik speaker and a linguist independently reviewed each utterance to verify that the recorded audio corresponded precisely with its transcript.

Particular attention was given to tonal realization and pronunciation accuracy, given the tonal nature of Efik. Utterances containing tonal inconsistencies, articulation errors, or misalignments between speech and text were flagged during review. Such instances were either corrected through re-transcription where appropriate or removed from the dataset to maintain high-quality paired text–audio alignment.

This validation process ensured that the final corpus meets the quality requirements for supervised neural TTS training, where even minor transcription or tonal errors can negatively affect acoustic modeling and intelligibility.

\subsection{Data Preprocessing}
All recordings were originally captured in .m4a format and subsequently converted to uncompressed .wav format to ensure compatibility with standard speech processing pipelines. The audio was resampled to 16kHz mono to maintain consistency across the corpus and align with common neural TTS training configurations.

Basic signal cleaning procedures were applied to improve acoustic consistency. Leading and trailing silence segments were trimmed while preserving natural prosodic boundaries. To prevent truncation of final phonemes, which is important for tonal realization, approximately 60-100 milliseconds of trailing silence were retained at the end of each utterance. This ensured that sentence-final acoustic cues were not clipped during segmentation.

Text preprocessing involved orthographic normalization to remove irrelevant punctuation marks and non-speech symbols. Sentence boundaries were preserved to maintain natural prosodic structure. All transcripts were manually verified for consistency with the spoken content.

The final corpus of 2,632 utterances was partitioned into training, validation and test sets as summarized in Table~\ref{tab:data_split}.

\begin{table}[!ht]
\centering
\begin{tabular}{|l|r|}
\hline
\textbf{Split} & \textbf{Utterances} \\
\hline
Train & 1,975 \\
Validation & 264 \\
Test & 393 \\
\hline
\end{tabular}
\caption{Dataset split statistics.}
\label{tab:data_split}
\end{table}

\subsection{Statistical Analysis of Dataset}
\begin{figure}[!ht]
  \centering
  \includegraphics[width=\linewidth]{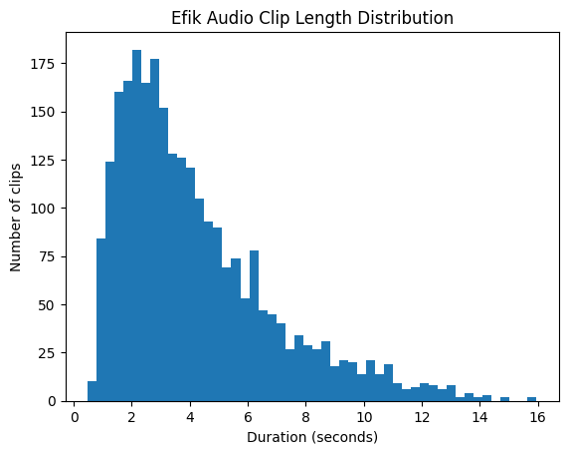}
  \caption{Distribution of audio clips duration.}
  \label{fig:audio_clips}
\end{figure}

The Efik TTS corpus consists of 2,632 utterances with a total duration of approximately 3.08 hours. The utterances vary in length from 0.48 seconds to 15.94 seconds, with an average duration of 4.21 seconds and a median duration of 3.49 seconds. Figure~\ref{fig:audio_clips} shows the distribution of audio clip durations, where the y-axis represents the number of clips and the x-axis represents duration in seconds.

\begin{figure}[!ht]
  \centering
  \includegraphics[width=\linewidth]{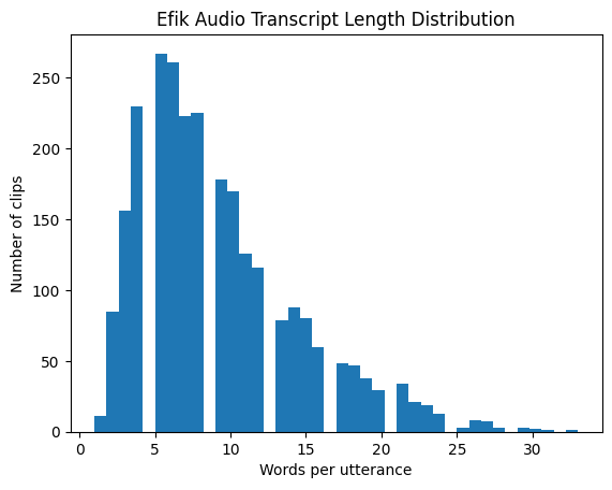}
  \caption{Distribution of transcript length per audio clip.}
  \label{fig:audio_transcripts}
\end{figure}

The corresponding text corpus contains 23,986 word tokens with a vocabulary of 3,276 unique words. Utterance lengths range from 1 to 33 words, with an average of 9.11 words per utterance. Figure~\ref{fig:audio_transcripts} illustrates the distribution of utterance lengths, with the y-axis representing the number of clips and the x-axis representing the number of words per utterance.

\section{TTS Models}
We fine-tuned four state-of-the-art (SOTA) Text-to-Speech (TTS) models for Efik: VITS, MMS-TTS, SpeechT5, and Orpheus-TTS. Each model was selected based on its ability to perform high-quality single-speaker TTS under low-resource conditions, which is critical for our dataset of 3 hours of single-speaker speech.

VITS (Variational Inference with Adversarial Learning for End-to-End TTS) \cite{kim2021vits} is a single-stage model that directly generates raw audio waveforms from text using a conditional variational autoencoder combined with normalizing flows and adversarial learning. 

MMS-TTS \cite{pratap2023scaling} is part of the Multilingual Speech Synthesis (MMS) framework, leveraging pretrained multilingual representations to improve synthesis quality in low-resource languages. Its architecture enables cross-lingual transfer, which is useful for Efik given the lack of large-scale TTS corpora.

SpeechT5 \cite{ao2022speecht5} is a transformer-based sequence-to-sequence TTS model that unifies speech and text representations. It allows flexible encoding of linguistic and acoustic features.

Orpheus-TTS \cite{orpheus2023tts} is a neural TTS model optimized for waveform fidelity and naturalness. It uses adversarial training techniques to improve the realism of generated audio.

\section{Experiments}

The four TTS models were fine-tuned with hyperparameters optimized for low-resource, single-speaker Efik TTS. VITS was trained for 50 epochs with a learning rate of 2e-4, a batch size of 4, and the Adam optimizer. MMS-TTS was trained for 50 epochs with a learning rate of 2e-5, a batch size of 16, and the AdamW optimizer, leveraging multilingual pretraining to improve performance in low-resource settings. SpeechT5 was trained with a smaller learning rate of 1e-5 and a batch size of 4, and was allowed to run up to 2,500 epochs with early stopping to ensure stable convergence under limited data conditions. A dropout rate of 0.1 was applied to reduce overfitting. Orpheus-TTS was trained for 50 epochs with a learning rate of 2e-5 and a batch size of 8. All models were trained on a single NVIDIA A100 GPU using mixed-precision training to improve computational efficiency and memory utilization. Early stopping based on validation loss was applied to mitigate overfitting given the limited size of the dataset. The single-speaker configuration provided acoustic consistency across recordings, which is beneficial for learning stable spectral and prosodic representations, particularly in a tonal language such as Efik.

Since MMS-TTS had no pretrained checkpoint for Efik, we initialized it using a Yoruba checkpoint and extended its vocabulary and embedding to include Efik-specific characters such as \d{o}\ and \~n. Similarly, VITS and SpeechT5 required updates to their embeddings and vocabularies to accommodate these characters, whereas Orpheus-TTS worked without modification. Despite these efforts, all models struggled to correctly pronounce words containing \~n\, highlighting the challenge of representing rare phonemes in extremely low-resource settings. In contrast, for the \d{o}\ character, MMS-TTS, SpeechT5, and Orpheus-TTS were able to generate intelligible pronunciations, likely benefiting from pretraining on similar phonemes in other languages. VITS, however, failed to produce intelligible speech for many Efik utterances, reflecting the sensitivity of its architecture to extremely low-resource conditions and the necessity for larger datasets to learn both waveform and prosody jointly.

When evaluating long-sequence generation, MMS-TTS outperformed all other models, generating continuous Efik audio of up to three minutes without hallucinating or producing nonsensical output. By comparison, VITS, Orpheus-TTS, and SpeechT5 frequently produced incoherent or meaningless speech after 20–30 seconds. Among these, Orpheus performed better than VITS and SpeechT5 in terms of intelligibility, although the generated speech retained a distinctly foreign accent resembling a European male speaker and often failed to preserve Efik tonal and cultural nuances.

These observations indicate that while current TTS architectures can generate intelligible Efik speech in short sequences, high-quality, larger datasets are essential for improving naturalness, tonal accuracy, and long-form generation. MMS-TTS shows the greatest promise for low-resource, single-speaker Efik synthesis, but all models would benefit from additional data to capture the unique phonetic, prosodic, and cultural characteristics of the language. With more extensive, high-quality recordings, it is feasible to develop a TTS system capable of producing natural Efik speech that preserves both tonal distinctions and pronunciation nuances.

\subsection{Evaluation}
The performance of the four TTS models was evaluated using three complementary metrics: Mean Opinion Score (MOS) \cite{viswanathan2005measuring}, Naturalness MOS (Nat-MOS), and Accent MOS (A-MOS). MOS measures the overall perceived naturalness of synthesized speech, Nat-MOS focuses specifically on how natural the speech sounds to native Efik listeners, and A-MOS evaluates the degree to which the synthesized speech preserves Efik accent and phonetic characteristics. Ratings were collected from 5 native Efik speakers, who were asked to rate short audio clips on a 1–5 scale (1 = completely unnatural, 5 = highly natural). Each model was evaluated on a subset of test utterances, and the mean and standard deviation of the ratings are reported in Table~\ref{tab:efik_tts_mos}.

\begin{table}[th]
\caption{MOS evaluation of Efik TTS models by native speakers. Higher MOS indicates more natural or accurate speech.}
\label{tab:efik_tts_mos}
\centering
\begin{tabular}{l c c c}
\toprule
\textbf{Model} & \textbf{MOS} & \textbf{Nat-MOS} & \textbf{A-MOS} \\
\midrule
VITS & 1.08 ± 0.27 & 1.04 ± 0.19 & - \\
SpeechT5 & 2.48 ± 0.49 & 1.88 ± 0.51 & 1.64 ± 0.48 \\
Orpheus-TTS & 3.08 ± 0.48 & 2.32 ± 0.46 & 2.21 ± 0.43 \\
\textbf{MMS-TTS} & \textbf{3.80 ± 0.63} & \textbf{3.60 ± 0.56} & \textbf{3.04 ± 0.52} \\
\bottomrule
\end{tabular}
\end{table}

From the results, MMS-TTS achieved the highest MOS, Nat-MOS, and A-MOS among the TTS models, producing the most natural and intelligible Efik speech. MMS-TTS also performed best in long-form generation, generating coherent speech close to 3 minutes without hallucinating, whereas other models often produced incoherent outputs after 20–30 seconds. VITS performed the worst across all metrics, likely due to its architectural reliance on large-scale data and sensitivity to extremely low-resource, single-speaker datasets. SpeechT5 achieved moderate scores, producing intelligible speech for short utterances but struggling with tonal accuracy, long-sequence generation, and accent preservation. Orpheus-TTS performed better than VITS and SpeechT5 for short sequences, though it retained a slight foreign (European) accent and did not fully preserve Efik tonal and cultural nuances.

In general, the evaluation demonstrates that while current end-to-end TTS models can generate intelligible Efik speech, high-quality, larger datasets and specialized low-resource training strategies will be critical for achieving more natural, culturally aligned, and tonally accurate speech synthesis in African languages like Efik.

\section{Results and Discussion}
The MOS evaluation in Table~\ref{tab:efik_tts_mos} highlights clear performance differences among the four TTS models for Efik speech synthesis. MMS-TTS achieved the highest MOS (3.80 ± 0.63), producing the most natural and intelligible speech. Its strong performance is likely due to multilingual pretraining, which enables the model to leverage cross-lingual phonetic knowledge, particularly for tonal phonemes like \d{o}, and to generate long-form audio sequences of up to three minutes without hallucination.

Orpheus-TTS achieved a moderate MOS (3.08 ± 0.48), outperforming VITS and SpeechT5 in short sequences. However, native speakers noted a slight foreign accent, and the model struggled to fully preserve Efik tonal and cultural characteristics. Despite these limitations, it maintained coherent output for medium-length sequences, making it a viable option for low-resource single-speaker TTS when tonal precision is less critical.

SpeechT5 produced intelligible short utterances but exhibited hallucinations and incoherent output after 20–30 seconds. Its MOS of 2.48 ± 0.49 reflects both the issues and challenges in accurately reproducing tonal contours, which are essential for lexical and grammatical meaning in Efik.

VITS performed the worst, with an MOS of 1.08 ± 0.27. The model struggled to generate intelligible audio and reliably capture tonal variations. This outcome is expected given VITS’s reliance on large-scale datasets for training; the 3-hour single-speaker dataset was insufficient. Consequently, longer utterances were often incoherent and unnatural.

This study shows that current state-of-the-art TTS architectures can produce intelligible Efik speech, but further work is required to achieve natural, culturally aligned, and tonally accurate synthesis. Future directions include expanding the dataset, incorporating multiple speakers to enrich prosody, and exploring architecture modifications or phoneme-level conditioning to better handle tonal and rare phonemes.

\section{Conclusion}
This work presents the first systematic effort to develop Text-to-Speech systems for Efik, a low-resource tonal African language. We fine-tuned four state-of-the-art end-to-end TTS models, VITS, MMS-TTS, SpeechT5, and Orpheus-TTS, using a single-speaker, three-hour dataset. Evaluation with native speakers shows that MMS-TTS achieved the highest naturalness, particularly for long-form generation, while VITS struggled due to its architectural reliance on large datasets. Orpheus-TTS and SpeechT5 produced intelligible short sequences but were less successful in preserving tonal and culturally aligned speech patterns.

While prior work on African languages such as Yoruba, Hausa, and Swahili \cite{ogunremi2024iroyinspeech, aliero2020crosslingual, mbonimpa2025edge} has laid the groundwork for TTS in low-resource settings, Efik presents unique challenges due to its extremely limited data and tonal complexity. These findings indicate that achieving truly natural, culturally aligned, and tonally accurate Efik speech synthesis requires larger, high-quality datasets, careful phoneme coverage, and model adaptations tailored for low-resource tonal languages. This study builds on existing African language TTS research and provides a foundation for future work in Efik and other underrepresented African languages, highlighting the critical role of speech technologies in digital language preservation.

\section{Limitations}
This study is limited by the use of a single-speaker dataset and a total of only 3 hours of audio, which constrained prosodic variation and long-sequence modeling, particularly for models like VITS. Rare phonemes such as \~n\ posed challenges across all models, and while MMS-TTS handled tonal patterns reasonably well, Orpheus-TTS and SpeechT5 produced speech with slight foreign accents, failing to fully capture Efik tonal and cultural nuances. Overcoming these limitations will require larger, multi-speaker datasets and specialized training strategies designed for low-resource tonal languages.

\section{Ethics Statement}
The single native Efik speaker provided informed consent for the recording and use of their voice for research purposes in text-to-speech development. All collected materials, including narrative and educational texts, were used in accordance with copyright and cultural guidelines.

We recognize the potential risks associated with speech synthesis systems, particularly the possibility of voice misuse, such as generating speech that could impersonate the recorded speaker without authorization. To mitigate this, the dataset and resulting models are intended strictly for non-commercial research use and will be released under a non-commercial license.

Access to the dataset and trained models will be restricted to researchers who agree to the license terms, which explicitly prohibit malicious or deceptive use, including impersonation or fraudulent content generation.

We will also document these risks clearly to raise awareness of responsible use of low-resource speech synthesis systems. No sensitive personal data is included in this work.

\section{Acknowledgements}
We sincerely thank the native Efik speakers for their invaluable contribution, and the linguist who carefully validated the dataset. We also extend our appreciation to Dr. David Adelani, Luel Hagos, Saheed Azeez, Abraham Owodunni, Gideon george and Steven Kolawole for their guidance, insightful advice, and support throughout this project. We especially appreciate the Machine Learning Collective (ML Collective) community for their unwavering support and guidance toward the success of this research work.

\section{Generative AI Use Disclosure}
No generative AI tools were used to generate scientific content, analyses, results, figures, or conclusions in this manuscript. Any use of generative AI was limited to grammar correction and language editing.

\bibliographystyle{IEEEtran}
\bibliography{mybib}

\end{document}